\documentclass[runningheads,a4paper]{llncs}

\usepackage{amsmath}
\usepackage{amssymb}
\usepackage{graphicx}
\usepackage{cite}
\usepackage{url}
\usepackage{color}
\usepackage{wrapfig}
\urldef{\mailsa}\path|{olsonran@upenn.edu}

\newcommand{\keywords}[1]{\par\addvspace\baselineskip
\noindent\keywordname\enspace\ignorespaces#1}

\newcommand{\be}{\begin{eqnarray}}
\newcommand{\ee}{\end{eqnarray}}
\begin{document}

\mainmatter  

\title{Evolution of active categorical image classification via saccadic eye movement}

\titlerunning{Evolution of active categorical image classification}

%
%
\author{Randal S.~Olson\inst{1}
\and Jason H.~Moore\inst{1}
\and Christoph Adami\inst{2}}
\authorrunning{Randal S.~Olson et al.} 

\institute{Institute for Biomedical Informatics, University of Pennsylvania\\
\email{olsonran@upenn.edu, jhmoore@upenn.edu}
\and
Department of Microbiology and Molecular Genetics, Michigan State University\\
\email{adami@msu.edu}}


%
%

\maketitle

\begin{abstract}

Pattern recognition and classification is a central concern for modern information processing systems. In particular, one key challenge to image and video classification has been that the computational cost of image processing scales linearly with the number of pixels in the image or video. Here we present an intelligent machine (the ``active categorical classifier,'' or ACC) that is inspired by the saccadic movements of the eye, and is capable of classifying images by selectively scanning only a portion of the image. We harness evolutionary computation to optimize the ACC on the MNIST hand-written digit classification task, and provide a proof-of-concept that the ACC works on noisy multi-class data. We further analyze the ACC and demonstrate its ability to classify images after viewing only a fraction of the pixels, and provide insight on future research paths to further improve upon the ACC presented here.

\end{abstract}

\section{Introduction}

Pattern recognition and classification is one of the most challenging ongoing problems in computer science in  which we seek to classify objects within an image into categories, typically with considerable variation among the objects within each category. With {\em invariant} pattern recognition, we seek to develop a model of each category that captures the essence of the class while compressing inessential variations. In this manner, invariant pattern recognition can tolerate (sometimes drastic) variations within a class, while at the same time recognizing differences across classes that can be minute but salient. One means of achieving this goal is through invariant feature extraction~\cite{Trieretal1996}, where the image is transformed into feature vectors that may be invariant with respect to a set of transformations, such as displacement, rotation, scaling, skewing, and lighting changes. This method can also be used in a hierarchical setting, where subsequent layers extract compound features from features already extracted in lower levels, such that the last layer extracts features that are essentially the classes themselves~\cite{LeCun1989}. Most of these existing methods have one thing in common: they achieve invariance either by applying transformations to the image when searching for the best match, or by mapping the image to a representation that is itself invariant to such transformations.

In contrast to these ``passive'' methods where transformations are applied to the image, we propose an active, attention-based method, where a virtual camera roams over and focuses on particular portions of the image, similar to how our own brain controls the focus of our attention~\cite{Mnih2014}. In this case, the camera's actions are guided by what the camera finds in the image itself: In essence, the camera searches the image to discover features that it recognizes, creating in the process a time series of experiences that guides further movements and eventually allows the camera to classify the image. We call this camera an ``active categorical classifier,'' or ACC for short.

Broadly speaking, the problem of classifying a spatial pattern is transformed into one of detecting differences within and between time series, namely the temporal sequence that the virtual camera generates in its sensors as it navigates the image. The method we propose here is inspired by models of visual attention~\cite{IttiKoch2001}, where attention to ``salient'' elements of an image or scene is guided by the image itself, such that only a small part of the incoming sensory information reaches short-term memory and visual awareness. Thus, focused attention overcomes the information-processing bottleneck imposed by massive sensory input (which can easily be $10^7-10^8$ bits per second in parallel at the optic nerve~\cite{IttiKoch2001}), and serializes this stream to achieve near-real-time processing with limited computational requirements.

In previous work, we have shown that it is possible to evolve robust controllers that navigate arbitrary mazes with near-perfect accuracy~\cite{Edlund2011} and simulate realistic animal behavior~\cite{Olson2013PredatorConfusion}. Independently, we have shown that we can evolve simple spatial classifiers for hand-written numerals in the MNIST data set~\cite{Chapmanetal2013}. Here we use the same technology to evolve active categorical classifiers that ``forage'' on images and respond to queries about what they saw in the image without needing to examine the image again.


\section{Methods}

In this section, we describe the methods used to evolve the active categorical classifiers (ACCs). We begin by describing the simulation environment in which the ACC scans and classifies the images. Next, we outline the structure and underlying neural architecture of an ACC. Finally, we provide details on the evolutionary process that we used to evolve the ACCs and the experiments that we conducted to evaluate them.

\subsection{Simulation Environment}

We evaluate the ACC on the MNIST data set, which is a well-known set of hand-written digits commonly used in supervised image classification research~\cite{LeCunetal1998}. The MNIST data set contains 28x28 pixel images of hand-written digits---all with corresponding labels indicating what digit the image represents (0--9)---and comes in two predefined sets of training and testing data (60,000 and 10,000 images, respectively). In this project, we binarize the images such that any pixels with a grayscale value $>127$ (out of the range [0, 255]) are assigned a value of 1, and all other pixels are assigned a value of 0.

When we evaluate an ACC, we place it at a random starting point in the 28x28 image and provide it a maximum of 40 steps to scan the image and assign a classification. (The 40-step maximum is meant to limit each simulation to a reasonably short amount of time.) Every simulation step, the ACC decides 1) what direction to move, 2) what class(es) it currently classifies the image as, and 3) whether it has made its final classification and is ready to terminate the simulation early. The ACC is evaluated only on its final classification for each image in the training set, with a ``fitness'' score ($F_{\rm ind}$) assigned as:
\begin{equation}
F_{\rm ind} = \frac{1}{1000} \times \sum_{i=1}^{1000} \frac{\rm CorrectClass_i}{\rm NumClassesGuessed_i}
\label{eq:fitness}
\end{equation}
where $i$ is the index of an individual image in the training set, ${\rm CorrectClass}_i=1$ if the correct class is among the ${\rm NumClassesGuessed}_i$ guesses that the ACC offers (it is allowed to guess more than one), and ${\rm CorrectClass}_i=0$ otherwise. Thus, an ACC can achieve a minimum fitness of 0.1 by guessing {\em all} classes for all images, but only achieves a maximum fitness of 1.0 by guessing the correct class only for every image. We note that due to computational limitations, we subset the MNIST training set to the first 100 images of each digit, such that we use only 1,000 training images in total (1/60th of the total set).

\begin{figure}[hb]
\begin{center}
\includegraphics[width=2in]{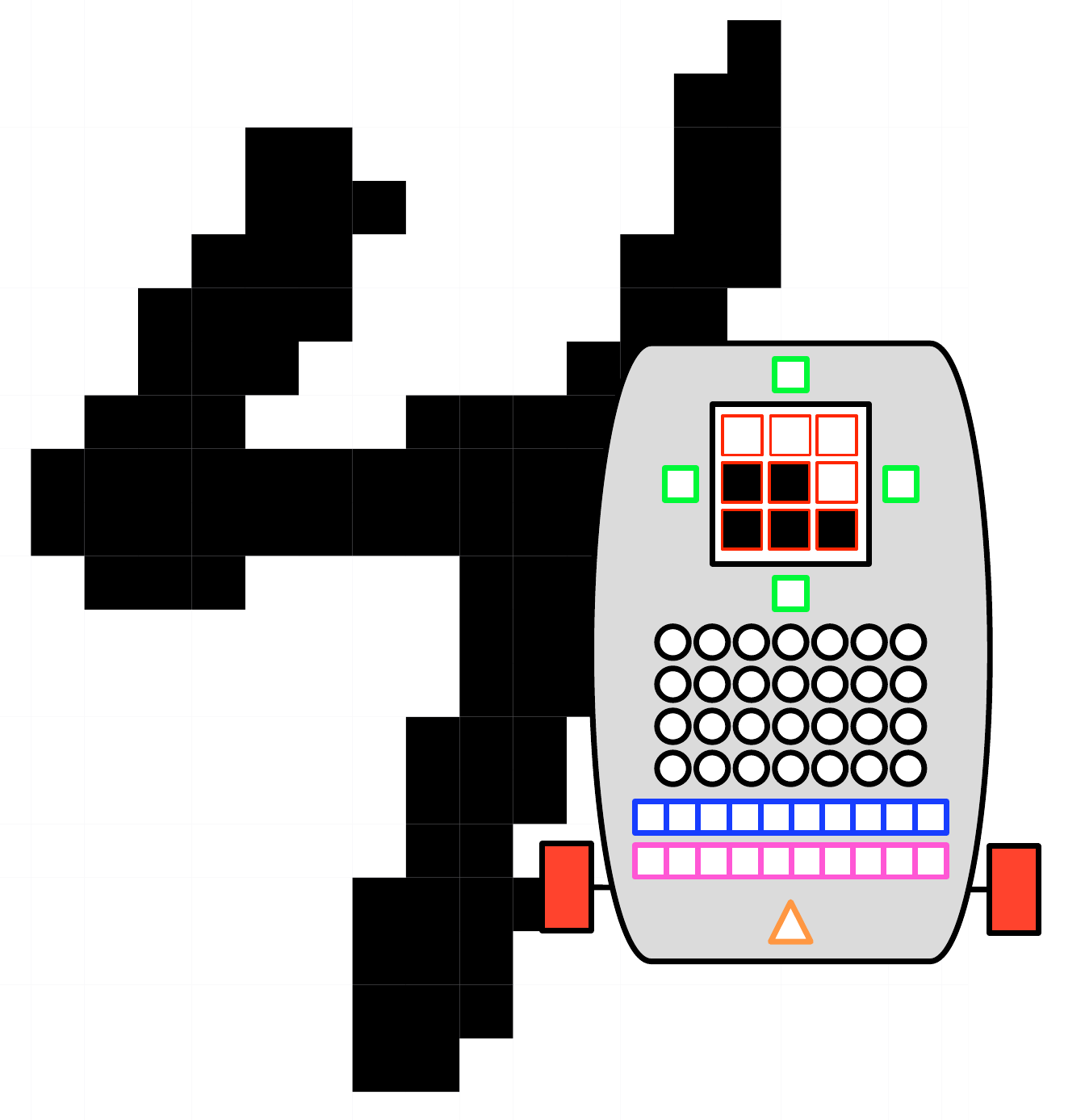}
\end{center}
\caption{
{\bf Active categorical classifier (ACC) configuration.} The ACC brain has 64 binary states that either fire or are quiescent, and represent sensory input from the image, internal memory, or decisions about how to interact with the image (described in the text).}
\label{fig:acc-config}
\end{figure}

\subsection{Active Categorical Classifier (ACC)}

We show in Fig.~\ref{fig:acc-config} the ACC in its natural habitat, roaming a digitized MNIST numeral. Each ACC has a brain that consists of 64 Markov neurons (``states'') that either fire (state = 1) or are quiescent (state = 0), and represent sensory input from the image, internal memory, and decisions about how to interact with the image. The ACC uses nine of these states to view nine pixels of the image in a 3x3 square, and four of the states to probe for activated pixels outside of its field of view with four raycast sensors that project across the image from the 0$^{\circ}$, 90$^{\circ}$, 180$^{\circ}$, and 270$^{\circ}$ angles of the 3x3 square (green squares in Fig.~\ref{fig:acc-config}). The raycast sensors activate only when they intersect with an activated pixel, and allow the ACC to find the numeral even if its starting position is far from it.

We also provide the ACC two actuator states (``motor neurons'') that allow it to ``saccade'' three pixels up/down and left/right, or any combination thereof (red rectangles denoted as wheels in Fig.~\ref{fig:acc-config}). In addition, the ACC has 20 states dedicated to classifying the image: 10 states that can be activated to guess each digit class (blue squares), and 10 states to {\em veto} an activated guess for each digit class (purple squares), e.g., ``this is definitely not a 4.'' This configuration allows the ACC to guess multiple classes at once, and combine its internal logic to veto any of those guesses if it believes them to be incorrect. Finally, the ACC has a ``done'' state (orange triangle), which allows it to end the simulation early if it has already decided on its final guess(es) for the current image. The remaining 28 neurons are ``memory'' states (black circles) used to process and store information, and integrate that information over time.

The ``artificial brain'' for the ACC in these experiments is a {\em Markov Network} (MN, see, e.g.,~\cite{Edlund2011,Marstalleretal2013,Chapmanetal2013}) that deterministically maps the 64 states (described above) at time $t$ to a corresponding series of output states that we interpret to determine the ACC's movement actions and classifications at time $t + 1$. The combination of output states and sensory inputs from time $t + 1$ are then used to determine the output states for the ACC at time $t + 2$, and so on. Every MN must therefore usefully combine the information provided over time in the 64 states to decide where to move, classify the image, and finally to decide when it has gathered enough information to make an accurate classification. Making all these decisions at once requires complex logic that is difficult to design.

\subsection{Optimization Process}

In order to create the complex logic embodied by a Markov Network, we {\em evolve} the MNs to maximize classification accuracy on the training images. We use a standard Genetic Algorithm (GA) to stochastically optimize a population of byte strings~\cite{Eiben2003}, which deterministically map to the MNs that function as the ACC's ``artificial brains'' in the simulation described above. Due to space limitations, we cannot describe MNs in full detail here; a detailed description of MNs and how they are evolved can be found in~\cite{Olson2016SelfishHerd}.

In our experiments, the GA maintains a population of 100 byte strings (``candidates'') of variable length (maximum = 10,000 bytes) and evaluates them according to the fitness function in Equation~\ref{eq:fitness}. The GA selects the candidates to reproduce into the next generation's population via tournament selection, where it shuffles the population and competes every byte string against only one other byte string. In each tournament, the byte string with the highest fitness produces one exact copy of itself as well as one mutated copy of itself into the next generation, while the ``loser'' produces no offspring. We note that the GA applies only mutations to the offspring (no crossover/recombination), with a per-byte mutation rate of 0.05\%, a gene duplication rate of 5\%, and a gene deletion rate of 2\%.

\subsection{Experiments}

According to the evolutionary optimization process, the GA selects ACCs that are capable of spatio-temporal classification of MNIST digits. We first ran 30 replicates of the GA with random starting populations and distinct random seeds and allowed these replicates to run for 168 hours on a high-performance compute cluster. From those 30 replicates, we identified the highest-fitness ACC (the ``elite''), and seeded another set of 30 replicates with mutants of the elite ACC. We allowed this second set of replicates to run for another 168 hours. In the following section, we report on the results of these experiments.

\begin{figure}[hb]
\begin{center}
\includegraphics[width=0.9\textwidth]{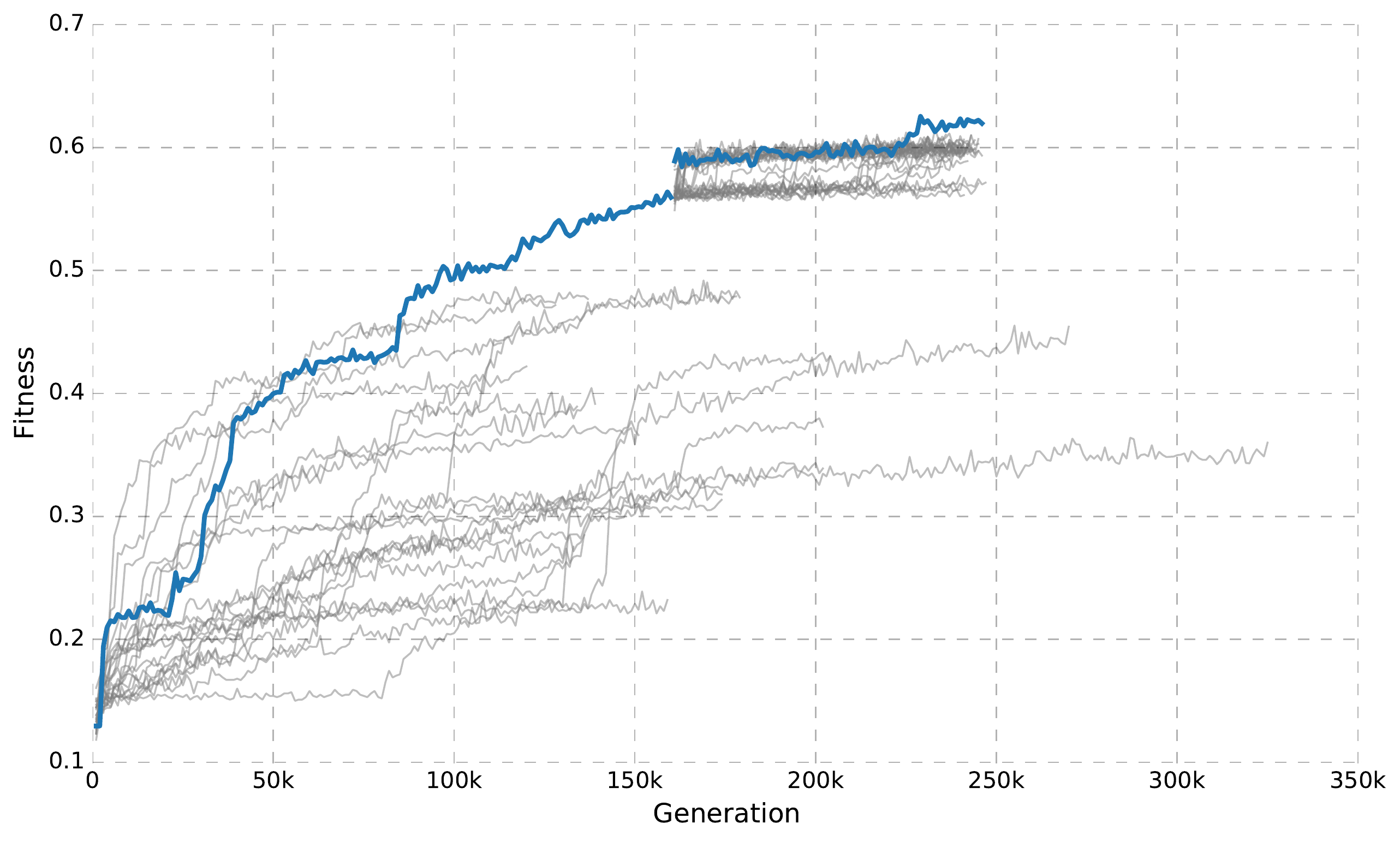}
\end{center}
\caption{
{\bf Fitness over time on the MNIST training set.} Each line represents a replicate of the evolutionary process that trains the active categorical classifiers. The lines represent the highest-fitness individual every 1,000 generations, where the blue line traces the lineage that led to the highest-fitness individual out of all replicates. After running all 30 replicates for one week, we took the best individual from the first set of runs and seeded another set of evolutionary runs with it, which is represented by the cluster of lines following the top lineage of first set.}
\label{fig:edd-mnist-fitness-over-time}
\end{figure}

\begin{figure}[t]
\begin{center}
\includegraphics[width=0.9\textwidth]{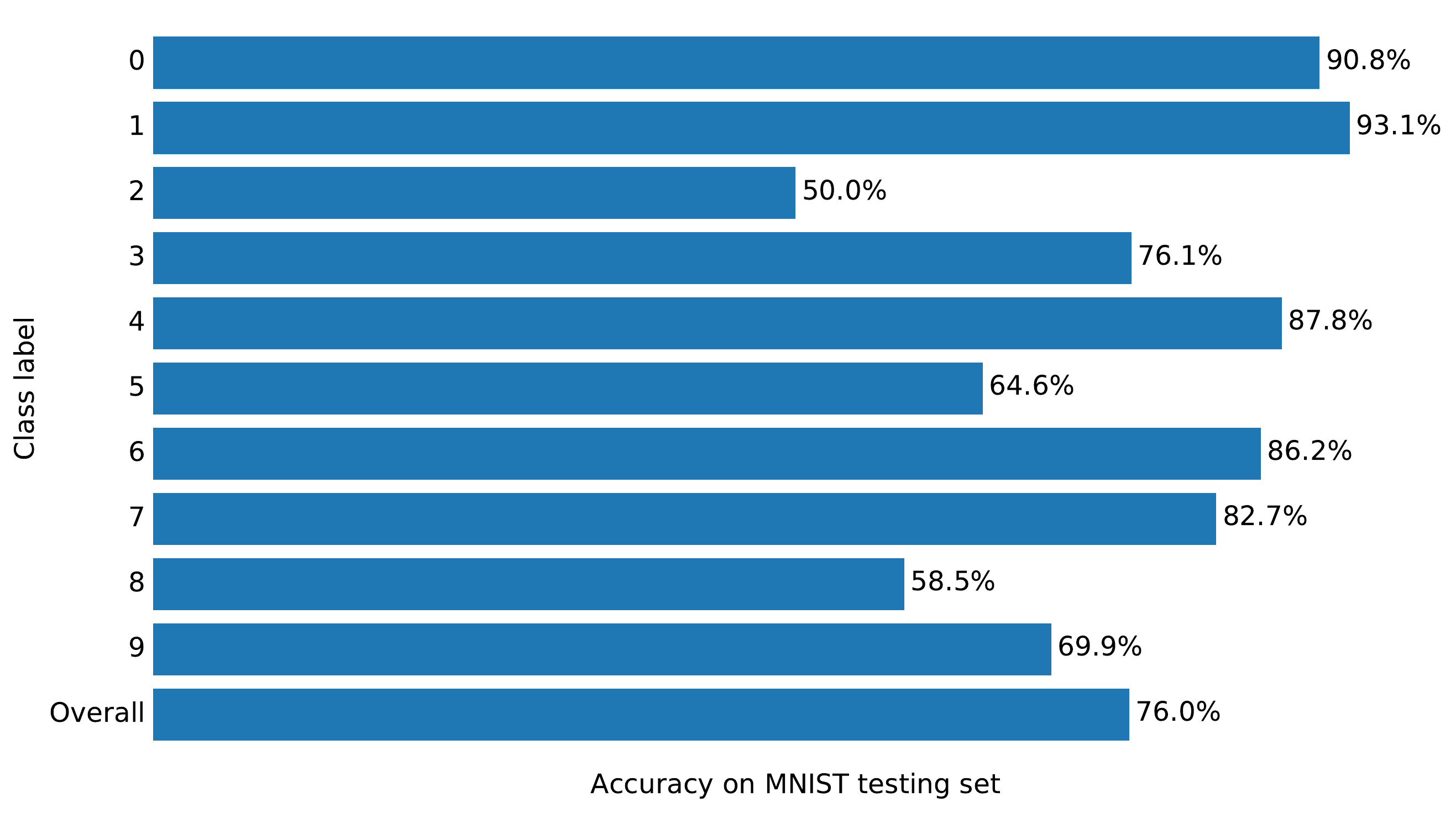}
\end{center}
\caption{
{\bf Active categorical classifier (ACC) accuracy on the binarized MNIST testing set.} We report per-digit accuracy (labeled 0--9) of the ACC as well as the average accuracy across all digits (labeled ``Overall'').}
\label{fig:edd-mnist-accuracy}
\end{figure}

\section{Results}

At the completion of the second set of replicates, the remaining active categorical classifiers (ACCs) had been optimized for 336 hours and roughly 250,000 generations. Shown in Fig.~\ref{fig:edd-mnist-fitness-over-time}, the ACCs experienced the majority of their improvements within the first 150,000 generations, and minimal improvements occurred in the second set of replicates, indicating that the ACCs had reached a plateau---either because the scan pattern required to improve was too complex, or because improving the classification accuracy on poorly classified digits compromised the ability to classify those digits the ACC was already proficient at. Such trade-offs are likely due to insufficient brain size, and investigations with larger brains are currently underway.

Instead of continuing the optimization process for a third set of replicates, we identified the highest-fitness ACC from replicate set 2 (highlighted in blue, Fig.~\ref{fig:edd-mnist-fitness-over-time}) and analyzed its spatio-temporal classification behavior to gain insights into its functionality. For the remainder of this section, we focus on the best ACC evolved in replicate set 2, which we will simply call ``the ACC.'' Shown in Fig.~\ref{fig:edd-mnist-accuracy}, the ACC achieved respectable but not state-of-the-art performance on the MNIST testing set: It managed to classify most of the 0s and 1s correctly for example, but failed to classify many of the 2s. Overall, the ACC achieved a macro-averaged accuracy of 76\%, which provides a proof-of-concept that the ACC works, but still has room for improvement on noisy multi-class data sets. We note that we have optimized ACCs on a set of hand-designed, non-noisy digits, where they managed to achieve 100\% accuracy. Thus, it is clear that the ACC architecture requires additional experimentation to fully adapt to noisy data, much like other methods currently in use. 
\begin{figure}[t]
\begin{center}
\includegraphics[width=0.9\textwidth]{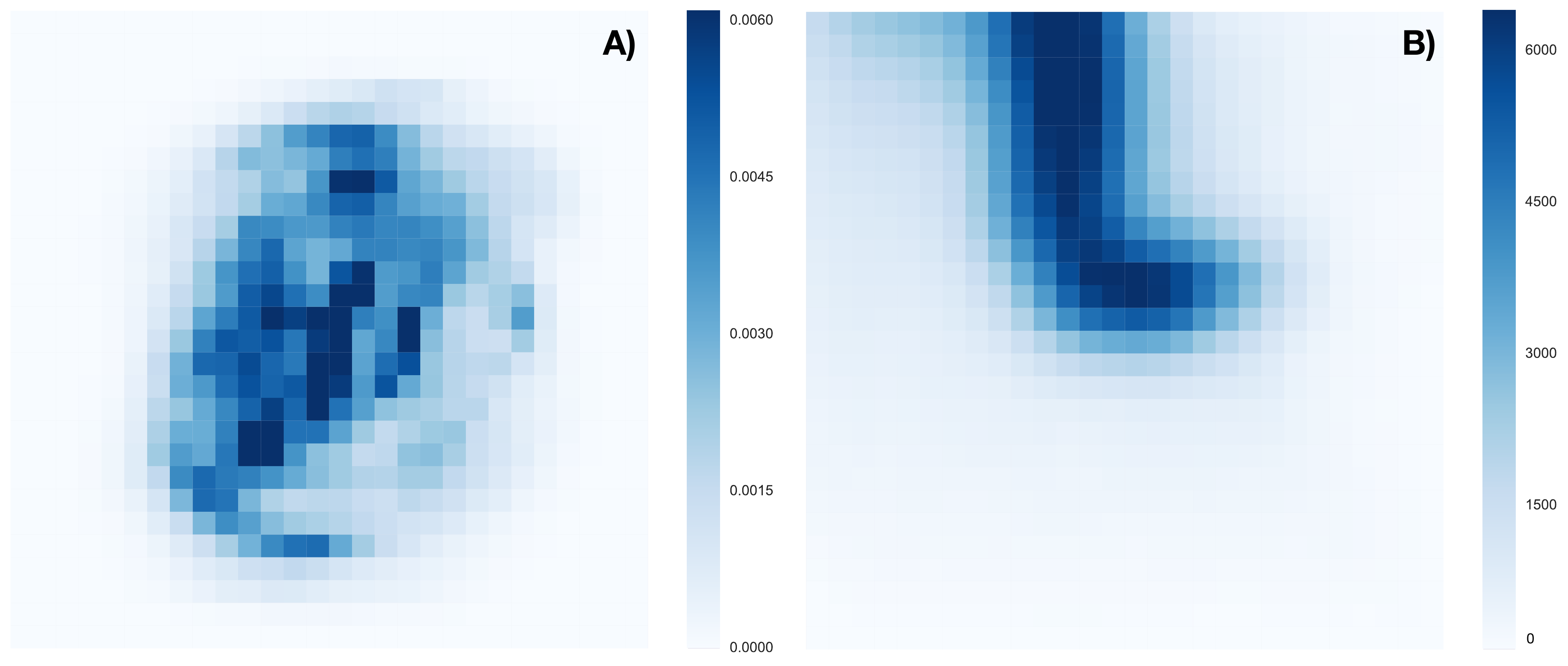}
\end{center}
\caption{
{\bf Analysis of informative pixels in the MNIST training set.} Panel A shows the most informative pixels in the MNIST training set according to feature importance scores from a Random Forest (i.e., Gini importance~\cite{GiniImportance}), whereas Panel B shows the pixels that the best active categorical classifier visited most frequently when classifying the MNIST data set. In both cases, darker colors represent higher values.}
\label{fig:edd-mnist-pixel-analysis}
\end{figure}
\begin{figure}[t]
\begin{center}
\includegraphics[width=0.75\textwidth]{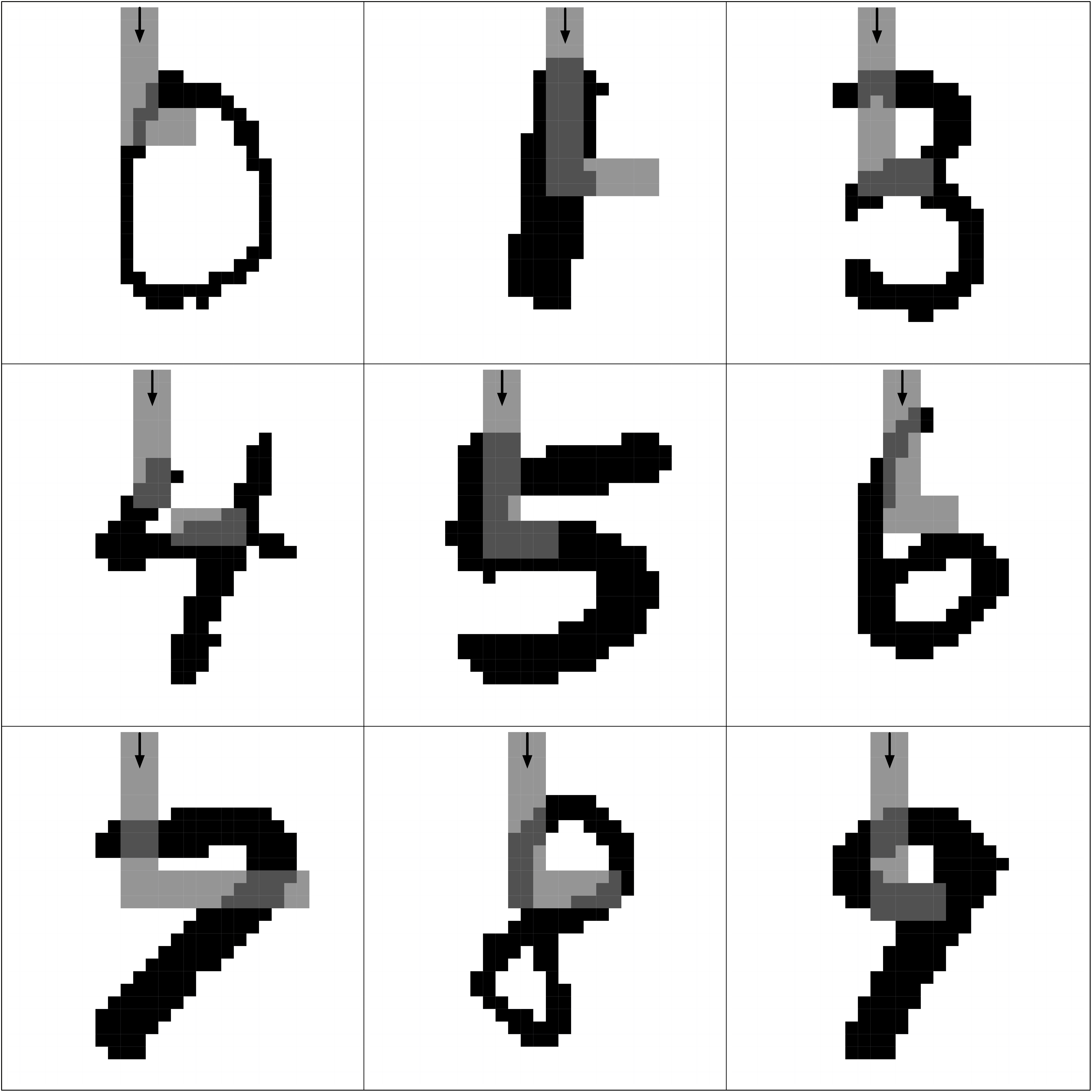}
\end{center}
\caption{
{\bf Example trajectories of the best active categorical classifier (ACC).} The arrows indicate the direction that the ACC followed, whereas the dark grey areas indicate the pixels that it scanned. Although the ACC starts all evaluations at random spots in the grid, it aligns itself to the digit to a common starting point and executes and L-shaped scan of the digit. We note that we excluded an example of digit 2 because the ACC rarely classifies it correctly, although it follows a similar L-shaped trajectory.}
\label{fig:edd-mnist-agent-static}
\end{figure}
In Fig.~\ref{fig:edd-mnist-pixel-analysis}B, we analyze the movement patterns of the ACC by counting how many times each pixel is viewed in the ACC's 3x3 visual grid when classifying the MNIST data set. Even though the ACC always starts at a random location in the image, we find that it follows a stereotypical scanning patterns of the digits: the ACC lines itself up to the top-left of the digit, then executes an L-shaped scanning pattern.

In contrast, Fig.~\ref{fig:edd-mnist-pixel-analysis}A depicts the most informative pixels for differentiating the classes in the binarized MNIST data set with a Random Forest classifier as implemented in scikit-learn~\cite{scikit-learn}. Here, we find that the most informative pixels exist in the center of the images, with several less-informative pixels on the image edges. Importantly, we note that the ACC never scans some of the most informative pixels in the lower half of the MNIST images (Fig.~\ref{fig:edd-mnist-pixel-analysis}A vs. Fig.~\ref{fig:edd-mnist-pixel-analysis}B). We believe that this behavior is the reason that the ACC is rarely able to classify any of the 2s, for example, because some of the most critical pixels for differentiating 2s from the rest of the digits are never visited.

We provide examples of the ACC scanning patterns in Fig.~\ref{fig:edd-mnist-agent-static}. Shown again is the stereotypical L-shaped scanning pattern starting at the upper-left corner of every digit. (We note that we trimmed the agent paths to only the final scanning pattern because the initial phase of ACC movements are simply lining up to the upper-left corner of the digit.) Interestingly, the ACC scans only a fraction of the available pixels to make each classification, and appears to be integrating information about the digit over space and time to identify distinctive sub-features of the digits. Furthermore, the ACC completes the majority of its scans within 5--10 steps and then immediately activates the ``done'' state, indicating that the ACC also learned when it knows the correct digit.

\section{Discussion}

The results that we display here show that it is possible to optimize an active categorical classifier (ACC) that scans a small portion of an image, integrates that information over space and time, and proceeds to perform an accurate classification of the image. Although the ACC does not achieve competitive accuracy on the MNIST data set compared to many modern techniques (76\% testing accuracy, Fig.~\ref{fig:edd-mnist-accuracy}), we believe that this result is due to the lack of training data rather than any particular limitation of ACCs: Due to computational limitations, we were only able to use a fixed set of 1,000 training images (100 of each class) to optimize the ACCs, while modern techniques use much larger training sets that even include additional variations of the training images~\cite{Wan2013}. Indeed, when we trained a scikit-learn Random Forest with 500 decision trees~\cite{scikit-learn} on the same binarized training set of 1,000 images, it achieves only 88.5\% accuracy on the MNIST testing set as compared to 97.5\% when it is trained on the full training set. Thus, in future work we will focus on integrating methods that expose the ACCs to all training images in an efficient manner.

From the point of view of embodied artificial intelligence, the challenge presented to the ACC in the image classification task is remarkably difficult. For one, these experiments challenged a single artificial brain to simultaneously perform several complex tasks, including to line itself up to a consistent starting point regardless of where it randomly starts in the image, decide where it needs to move to complete the scan based on limited information about the image, determine what pixels are important to consider, {\em and} integrate that information over space and time to classify the image into 1 of 10 classes. We furthermore challenged the ACC to evolve something akin to a ``theory of mind'' such that it knows when it has guessed the correct class for the image and to end the simulation early. In future work, it will be illuminating to analyze the underlying neural architecture of the evolved ACCs to provide insight into the fundamentals of active categorical perception~\cite{Beer2003}.

Unlike many modern image classification techniques that must analyze an entire static image to determine an image's class, the ACC instead integrates information from a small subset of the pixels over space and time. This method naturally lends itself to video classification, where feature compression will play a crucial role in overcoming the massive data size challenge for real-time classification of moving objects~\cite{Mnih2014}.
Lastly, recent work has shown that modern deep learning-based image classification techniques tend to be easily fooled because they are trained in a supervised, discriminative manner: They establish decision boundaries that appropriately separate the data they encounter in the training phase, but these decision boundaries also include (and thus mis-classify) many inappropriate data points never encountered during training~\cite{Nguyen2015}. Although most deep learning researchers respond to this challenge by creating additional ``adversarial'' training images to train the deep neural networks~\cite{Goodfellow2014}, we believe that the findings in~\cite{Nguyen2015} highlight a critical weakness in deep learning: the resulting networks are trained to precisely map inputs to corresponding target outputs, without generalizing far beyond the training data they are exposed to~\cite{Szegedy2013}.

Due to their nature, deep neural networks are highly dependent on the training data, and only generalize to new challenges if they are similar to those encountered in the training data~\cite{Goodfellow2014}. In contrast, heuristic-based machines such as the ACC learn simple, generalizable heuristics for classifying images that encode the conceptual representation~\cite{Marstalleretal2013} of the objects, and should not be so easily fooled. As such, even if the ACC in the present work does not achieve competitive accuracy when compared to modern deep learning techniques, we believe that further development of heuristic-based image classification machines will lead to robust classifiers that will eventually surpass deep neural networks in generalizability without the need for adversarial training images. We further believe that it is precisely those machines that carry with them complex representations of the world that will become the robust and sophisticated intelligent machines of the future. Whether the embodied evolutionary approach we describe here will succeed in this is, of course, an open problem.

\section{Acknowledgments}

We thank David B.~Knoester, Arend Hintze, and Jeff Clune for their valuable input during the development of this project. We also thank the Michigan State University High Performance Computing Center for the use of their computing resources. This work was supported in part by the National Science Foundation BEACON Center under Cooperative Agreement DBI-0939454, and in part by National Institutes of Health grants LM009012, LM010098, and EY022300.

\bibliography{references}
\bibliographystyle{splncs}

\end{document}